\documentclass{article} % For LaTeX2e
\usepackage{nips14submit_e,times}
\usepackage{hyperref}
\usepackage{url}
\usepackage{amsmath}
\usepackage{amssymb}
\usepackage{mathrsfs}
\usepackage{amsthm}
\usepackage{color}
\newcommand{\bp}{\mathbf{p}}
\newcommand{\bq}{\mathbf{q}}

\newcommand{\bs}{\mathbf{s}}

\newcommand{\bx}{\mathbf{x}}

\newcommand{\bz}{\mathbf{z}}

\newcommand{\btheta}{\boldsymbol{\theta}}

\title{Distilling the Knowledge in a Neural Network}

\author{
Geoffrey Hinton\thanks{Also affiliated with the University of Toronto
  and the Canadian Institute for Advanced Research.}\ \ \thanks{Equal contribution.} \\
Google Inc.\\
Mountain View\\
\texttt{geoffhinton@google.com} \\
\And
Oriol Vinyals$^\dagger$\\
Google Inc.\\
Mountain View\\
\texttt{vinyals@google.com} \\
\And
Jeff Dean\\
Google Inc.\\
Mountain View\\
\texttt{jeff@google.com} \\
}

\nipsfinalcopy % Uncomment for camera-ready version

\begin{document}

\maketitle

\begin{abstract}

A very simple way to improve the performance of almost any machine
learning algorithm is to train many different models on the same data
and then to average their predictions \cite{Dietterich2000}.  Unfortunately, making
predictions using a whole ensemble of models is cumbersome and may be
too computationally expensive to allow deployment to a large number of
users, especially if the individual models are large neural
nets. Caruana and his collaborators \cite{Caruana} have shown that it
is possible to compress the knowledge in an ensemble into a single
model which is much easier to deploy and we develop this approach
further using a different compression technique. We achieve some
surprising results on MNIST and we show that we can significantly
improve the acoustic model of a heavily used commercial system by
distilling the knowledge in an ensemble of models into a single model. We
also introduce a new type of ensemble composed of one or more full
models and many specialist models which learn to distinguish
fine-grained classes that the full models confuse. Unlike a mixture of
experts, these specialist models can be trained rapidly and in
parallel.

\end{abstract}

\section{Introduction}

Many insects have a larval form that is optimized for extracting energy and nutrients from the environment and a
completely different adult form that is optimized for the very different requirements of traveling and reproduction.  In
large-scale machine learning, we typically use very similar models for the training stage and the deployment stage
despite their very different requirements: For tasks like speech and object recognition, training must extract structure
from very large, highly redundant datasets but it does not need to operate in real time and it can use a huge amount of
computation.  Deployment to a large number of users, however, has much more stringent requirements on latency and
computational resources.  The analogy with insects suggests that we should be willing to train very cumbersome models if
that makes it easier to extract structure from the data. The cumbersome model could be an ensemble of separately trained
models or a single very large model trained with a very strong regularizer such as dropout \cite{srivastava2014dropout}. Once the cumbersome model
has been trained, we can then use a different kind of training, which we call ``distillation'' to transfer the knowledge
from the cumbersome model to a small model that is more suitable for deployment.  A version of this strategy has
already been pioneered by Rich Caruana and his collaborators \cite{Caruana}. In their important paper they demonstrate
convincingly that the knowledge acquired by a large ensemble of models can be transferred to a single small model.

A conceptual block that may have prevented more investigation of this very promising approach is that we tend to
identify the knowledge in a trained model with the learned parameter values and this makes it hard to see how we can
change the form of the model but keep the same knowledge.  A more abstract view of the knowledge, that frees it from any
particular instantiation, is that it is a learned mapping from input vectors to output vectors.  For cumbersome models
that learn to discriminate between a large number of classes, the normal training objective is to maximize the average log
probability of the correct answer, but a side-effect of the learning is that the trained model assigns
probabilities to all of the incorrect answers and even when these probabilities are very small, some of them are much
larger than others.  The relative probabilities of incorrect answers tell us a lot about how the cumbersome model
tends to generalize.  An image of a BMW, for example, may only have a very small chance of being mistaken for a garbage
truck, but that mistake is still many times more probable than mistaking it for a carrot. 

It is generally accepted that the objective function used for training should reflect the true objective of the user as
closely as possible. Despite this, models are usually trained to optimize performance on the training data when the real
objective is to generalize well to new data. It would clearly be better to train models to generalize well, but this
requires information about the correct way to generalize and this information is not normally available.  When we are
distilling the knowledge from a large model into a small one, however, we can train the small model to generalize in the
same way as the large model. If the cumbersome model generalizes well because, for example, it is the average of a large
ensemble of different models, a small model trained to generalize in the same way will typically do much better on test
data than a small model that is trained in the normal way on the same training set as was used to train the ensemble. 

An obvious way to transfer the generalization ability of the cumbersome model to a small model is to use the class
probabilities produced by the cumbersome model as ``soft targets'' for training the small model. For this transfer
stage, we could use the same training set or a separate ``transfer'' set.  When the
cumbersome model is a large ensemble of simpler models, we can use an arithmetic or geometric mean of their individual predictive distributions as the soft targets.  When the soft
targets have high entropy, they provide much more information per training case than hard targets and much less variance
in the gradient between training cases, so the small model can often be trained on much less data than the original
cumbersome model and using a much higher learning rate.

For tasks like MNIST in which the cumbersome model almost always produces the correct answer with very high confidence,
much of the information about the learned function resides in the ratios of very small probabilities in the soft
targets. For example, one version of a 2 may be given a probability of $10^{-6}$ of being a 3 and $10^{-9}$ of being a 7
whereas for another version it may be the other way around. This is valuable information that defines a rich similarity
structure over the data ({\it i. e.} it says which 2's look like 3's and which look like 7's) but it has very little
influence on the cross-entropy cost function during the transfer stage because the probabilities are so close to zero.
Caruana and his collaborators circumvent this problem by using the logits (the inputs to the final softmax) rather than
the probabilities produced by the softmax as the targets for learning the small model and they minimize the squared
difference between the logits produced by the cumbersome model and the logits produced by the small model.  Our more
general solution, called ``distillation'', is to raise the temperature of the final softmax until the cumbersome model
produces a suitably soft set of targets. We then use the same high temperature when training the small model to match
these soft targets. We show later that matching the logits of the cumbersome model is actually a special case of
distillation.

The transfer set that is used to train the small model could consist entirely of unlabeled data \cite{Caruana} or we could use
the original training set.  We have found that using the original training set works well, especially if we add a small
term to the objective function that encourages the small model to predict the true targets as well as matching the soft
targets provided by the cumbersome model.  Typically, the small model cannot exactly match the soft targets and erring
in the direction of the correct answer turns out to be helpful.

\section{Distillation}\label{sec:distillation}

Neural networks typically produce class probabilities by using a ``softmax'' output layer that converts the logit,
$z_i$, computed for each class into a probability, $q_i$, by comparing $z_i$ with the other logits.
\begin{equation}
q_i = \frac{exp(z_i/T)}{\sum_j exp(z_j/T)}
\end{equation}
where $T$ is a temperature that is normally set to $1$.  Using a higher value for $T$ produces a softer probability
distribution over classes.

In the simplest form of distillation, knowledge is transferred to the distilled model by training it on a transfer set
and using a soft target distribution for each case in the transfer set that is produced by using the cumbersome model
with a high temperature in its softmax.  The same high temperature is used when training the distilled model, but after
it has been trained it uses a temperature of 1.

When the correct labels are known for all or some of the transfer set,
this method can be significantly improved by also training the
distilled model to produce the correct labels. One way to do this is
to use the correct labels to modify the soft targets, but we found
that a better way is to simply use a weighted average of two different
objective functions.  The first objective function is the cross
entropy with the soft targets and this cross entropy is computed using
the same high temperature in the softmax of the distilled model as was
used for generating the soft targets from the cumbersome model. The
second objective function is the cross entropy with the correct
labels. This is computed using exactly the same logits in softmax of
the distilled model but at a temperature of 1.  We found that the best
results were generally obtained by using a condiderably lower weight
on the second objective function. Since the magnitudes of the
gradients produced by the soft targets scale as $1/T^2$ it is
important to multiply them by $T^2$ when using both hard and soft
targets. This ensures that the relative contributions of the hard and
soft targets remain roughly unchanged if the temperature used for
distillation is changed while experimenting with meta-parameters.

\subsection{Matching logits is a special case of distillation}

Each case in the transfer set contributes a cross-entropy gradient, $dC/dz_i$, with respect to each logit, $z_i$ of the
distilled model.  If the cumbersome model has logits $v_i$ which produce soft target probabilities $p_i$ and the transfer training is done at a temperature of $T$,
this gradient is given by:
\begin{equation}
\frac{\partial C}{\partial z_i} = \frac{1}{T}\left(q_i - p_i\right) = \frac{1}{T}\left(\frac{e^{z_i/T}}{\sum_j e^{z_j/T}} -\frac{e^{v_i/T}}{\sum_j e^{v_j/T}}\right)
\label{deriv} 
\end{equation}

If the temperature is high compared with the magnitude of the logits, we can approximate:
\begin{equation}
\frac{\partial C}{\partial z_i} \approx \frac{1}{T}\left( \frac{1+z_i/T}{N + \sum_j z_j/T} - \frac{1+v_i/T}{N + \sum_j v_j/T} \right)
\label{mess}
\end{equation}

If we now assume that the logits have been zero-meaned separately for each transfer case so that $\sum_j z_j = \sum_j v_j = 0$ Eq. \ref{mess} simplifies to:
\begin{equation}
\frac{\partial C}{\partial z_i} \approx \frac{1}{NT^2}\left( z_i - v_i \right)
\label{nice}
\end{equation}
So in the high temperature limit, distillation is equivalent to
minimizing ${1/2}(z_i-v_i)^2$, provided the logits are zero-meaned
separately for each transfer case. At lower temperatures, distillation
pays much less attention to matching logits that are much more
negative than the average. This is potentially advantageous because
these logits are almost completely unconstrained by the cost function
used for training the cumbersome model so they could be very noisy.
On the other hand, the very negative logits may convey useful
information about the knowledge acquired by the cumbersome
model. Which of these effects dominates is an empirical question. We
show that when the distilled model is much too small to capture all of
the knowledege in the cumbersome model, intermediate temperatures work
best which strongly suggests that ignoring the large negative logits
can be helpful.

\section{Preliminary experiments on MNIST}

To see how well distillation works, we trained a single large neural
net with two hidden layers of 1200 rectified linear hidden units on
all 60,000 training cases. The net was strongly regularized using
dropout and weight-constraints as described in \cite{dropout}. Dropout
can be viewed as a way of training an exponentially large ensemble of
models that share weights. In addition, the input images were jittered
by up to two pixels in any direction.  This net achieved 67 test
errors whereas a smaller net with two hidden layers of 800 rectified
linear hidden units and no regularization achieved 146 errors. But if
the smaller net was regularized solely by adding the additional task
of matching the soft targets produced by the large net at a
temperature of 20, it achieved 74 test errors. This shows that soft
targets can transfer a great deal of knowledge to the distilled model,
including the knowledge about how to generalize that is learned from
translated training data even though the transfer set does not contain
any translations.

When the distilled net had 300 or more units in each of its two hidden
layers, all temperatures above 8 gave fairly similar results.  But
when this was radically reduced to 30 units per layer, temperatures in the range
2.5 to 4 worked significantly better than higher or lower
temperatures.

We then tried omitting all examples of the digit 3 from the transfer set.  So from the perspective of the distilled
model, 3 is a mythical digit that it has never seen. Despite this, the distilled model only makes 206 test errors of
which 133 are on the 1010 threes in the test set.  Most of the errors are caused by the fact that the learned bias for
the 3 class is much too low. If this bias is increased by 3.5 (which optimizes overall performance on the test set), the
distilled model makes 109 errors of which 14 are on 3s.  So with the right bias, the distilled model gets 98.6\% of
the test 3s correct despite never having seen a 3 during training. If the transfer set contains {\it only} the 7s and 8s from
the training set, the distilled model makes 47.3\% test errors, but when the biases for 7 and 8 are reduced by 7.6 to
optimize test performance, this falls to 13.2\% test errors.

\section{Experiments on speech recognition}\label{sec:speech}

In this section, we investigate the effects of ensembling Deep Neural Network (DNN) acoustic models that are used in
Automatic Speech Recognition (ASR).  We show that the distillation strategy that we propose in this paper achieves the
desired effect of distilling an ensemble of models into a single model that works significantly better than a model of
the same size that is learned directly from the same training data.

State-of-the-art ASR systems currently use DNNs to map a (short) temporal context of features derived
from the waveform to a probability distribution over the discrete states of a Hidden Markov Model (HMM) \cite{SPM}. More
specifically, the DNN produces a probability distribution over clusters of tri-phone states at each time and a decoder
then finds a path through the HMM states that is the best compromise between using high probability states and producing
a transcription that is probable under the language model.

Although it is possible (and desirable) to train the DNN in such a way that the decoder (and, thus, the language model)
is taken into account by marginalizing over all possible paths, it is common to train the DNN to perform frame-by-frame
classification by (locally) minimizing the cross entropy between the predictions made by the net and the labels given
by a forced alignment with the ground truth sequence of states for each observation:
$$\btheta = \arg\max_{\btheta'} P(h_t | \bs_t;\btheta')$$
where $\btheta$ are the parameters of our acoustic model $P$
which maps acoustic observations at time $t$, $\bs_t$, to a probability, $P(h_t | \bs_t;\btheta')$ , of the ``correct''
HMM state $h_t$, which is determined by a forced alignment with the correct sequence of words. The model is trained with
a distributed stochastic gradient descent approach. 

We use an architecture with 8 hidden layers each containing 2560 rectified linear units and a final softmax layer with
14,000 labels (HMM targets $h_t$).  The input is 26 frames of 40 Mel-scaled filterbank coefficients with a 10ms advance
per frame and we predict the HMM state of 21$^{st}$ frame.  The total number of parameters is about 85M. This is a
slightly outdated version of the acoustic model used by Android voice search, and should be considered as a very strong
baseline. To train the DNN acoustic model we use about 2000 hours of spoken English data, which yields about 700M
training examples. This system achieves a frame accuracy of 58.9\%, and a Word Error Rate (WER) of 10.9\% on our development set.

\subsection{Results}

We trained 10 separate models to predict $P(h_t | \bs_t;\btheta)$, using exactly the same architecture and training
procedure as the baseline. The models are randomly initialized with different initial parameter values and we find that
this creates sufficient diversity in the trained models to allow the averaged predictions of the ensemble to
significantly outperform the individual models.  We have explored adding diversity to the models by varying the sets of
data that each model sees, but we found this to not significantly change our results, so we opted for the simpler
approach. For the distillation we tried temperatures of $[1, {\bf 2}, 5, 10]$ and used a relative weight of 0.5 on the
cross-entropy for the hard targets, where bold font indicates the best value that was used for
table~\ref{tab:speech_results} .

Table~\ref{tab:speech_results} shows that, indeed, our distillation approach is able to extract more useful information
from the training set than simply using the hard labels to train a single model. More than 80\% of the improvement in
frame classification accuracy achieved by using an ensemble of 10 models is transferred to the distilled model which is
similar to the improvement we observed in our preliminary experiments on MNIST. The ensemble gives a smaller improvement
on the ultimate objective of WER (on a 23K-word test set) due to the mismatch in the objective function, but again, the
improvement in WER achieved by the ensemble is transferred to the distilled model.  

\begin{table}
\small
\centering
\begin{tabular}{|c|c|c|}
\hline
System & Test Frame Accuracy & WER \\
\hline
Baseline & 58.9\% & 10.9\%\\
10xEnsemble & 61.1\%& 10.7\% \\
Distilled Single model & 60.8\%& 10.7\%\\
\hline
\end{tabular}
\caption{Frame classification accuracy and WER showing that the distilled single model performs about as well as the
  averaged predictions of  10  models that were used to create the soft targets.}\label{tab:speech_results}
\end{table}

We have recently become aware of related work on learning a small acoustic model by matching the class probabilities of
an already trained larger model \cite{LiZhaoHuangGong}. However, they do the distillation at a temperature of 1 using a
large unlabeled dataset and their best distilled model only reduces the error rate of the small model by 28\% of the
gap between the error rates of the large and small models when they are both trained with hard labels.

\section{Training ensembles of specialists on very big datasets}

Training an ensemble of models is a very simple way to take advantage
of parallel computation and the usual objection that an ensemble
requires too much computation at test time can be dealt with by using
distillation. There is, however, another important objection to
ensembles: If the individual models are large neural networks and the
dataset is very large, the amount of computation required at training
time is excessive, even though it is easy to parallelize.

In this section we give an example of such a dataset and we show how
learning specialist models that each focus on a different confusable
subset of the classes can reduce the total amount of computation
required to learn an ensemble.  The main problem with specialists that
focus on making fine-grained distinctions is that they overfit very
easily and we describe how this overfitting may be prevented by using
soft targets.

\subsection{The JFT dataset}

JFT is an internal Google dataset that has 100 million labeled images
with 15,000 labels. When we did this work, Google's baseline model for
JFT was a deep convolutional neural network \cite{Kriz} that had been trained for
about six months using asynchronous stochastic gradient descent on a
large number of cores.  This training used two types of
parallelism \cite{brain-stuff}. First, there were many replicas of the neural net running
on different sets of cores and processing different mini-batches from
the training set. Each replica computes the average gradient on its
current mini-batch and sends this gradient to a sharded parameter server which
sends back new values for the parameters. These new values reflect all
of the gradients received by the parameter server since the last time
it sent parameters to the replica. Second, each replica is spread over
multiple cores by putting different subsets of the neurons on each
core. Ensemble training is yet a third type of parallelism that can be
wrapped around the other two types, but only if a lot more cores are
available. Waiting for several years to train an ensemble of models was
not an option, so we needed a much faster way to improve the baseline
model. 

\subsection{Specialist Models}\label{sec:specialists}

When the number of classes is very large, it makes sense for the
cumbersome model to be an ensemble that contains one generalist 
model trained on all the data and  many ``specialist''
models, each of which is trained on data that is highly enriched in
examples from a very confusable subset of the classes (like different
types of mushroom). The softmax of this type of specialist can be made
much smaller by combining all of the classes it does not care about into a
single dustbin class. 

To reduce overfitting and share the work of learning lower level feature detectors, each specialist model is initialized
with the weights of the generalist model. These weights are then slightly modified by training the
specialist with half its examples coming from its special subset and half sampled at random from the remainder of the
training set. After training, we can correct for the biased training set by incrementing the logit
of the dustbin class by the log of the proportion by which the specialist class is oversampled.

\subsection{Assigning classes to specialists}

In order to derive groupings of object categories for the specialists, we decided to focus on categories that our full
network often confuses. Even though we could have computed the confusion matrix and used it as a way to find such
clusters, we opted for a simpler approach that does not require the true labels to construct the clusters.

\begin{table}
\small
\centering
\begin{tabular}{|l|}
\hline
\textbf{JFT 1:} Tea party; Easter; Bridal shower; Baby shower; Easter Bunny;  ... \\
\textbf{JFT 2:} Bridge; Cable-stayed bridge; Suspension bridge; Viaduct; Chimney; ... \\
\textbf{JFT 3:} Toyota Corolla E100; Opel Signum; Opel Astra; Mazda Familia; ... \\
\hline
\end{tabular}
\caption{Example classes from clusters computed by our
  covariance matrix clustering algorithm}\label{tab:example_clusters}
\end{table}

In particular, we apply a clustering algorithm to the covariance matrix of the predictions of our generalist model, so that
a set of classes $S^m$ that are often predicted together will be used as targets for one of our specialist models, $m$. We applied an on-line
version of the K-means algorithm to the columns of the covariance matrix, and obtained reasonable clusters (shown in
Table \ref{tab:example_clusters}). We tried several clustering algorithms which produced similar results.

\subsection{Performing inference with ensembles of specialists}

Before investigating what happens when specialist models are
distilled, we wanted to see how well ensembles containing specialists
performed. In addition to the specialist models, we always have a
generalist model so that we can deal with classes for which we
have no specialists and so that we can decide which specialists to
use. Given an input image $\bx$, we do top-one classification in two steps:

Step 1: For each test case, we find
  the $n$ most probable classes according to the generalist model. Call this set of classes $k$. In our
  experiments, we used $n=1$.

Step 2: We then take all the specialist models, $m$, whose special
subset of confusable classes,
  $S^m$, has a non-empty intersection with $k$ and call this the active
  set of specialists $A_k$ (note that this set may be empty). We then
  find the full probability distribution $\bq$ over all the classes
  that minimizes:
\begin{equation}
KL (\bp^g, \bq) + \sum_{m \in A_k} KL (\bp^m , \bq)  
\label{eq:kl}
\end{equation}
where $KL$ denotes the KL divergence, and $\bp^m$ $\bp^g$ denote the
probability distribution of a specialist model or the generalist full
model. The distribution $\bp^m$ is a distribution over all the
specialist classes of $m$ plus a single dustbin class, so when
computing its KL divergence from the full $\bq$ distribution we sum
all of the probabilities that the full $\bq$ distribution assigns to
all the classes in $m$'s dustbin.

Eq.~\ref{eq:kl} does not have a general closed form solution, though when all the models produce a single probability for
each class the solution is either the arithmetic or geometric mean, depending on whether we use $KL(\bp,\bq)$
or $KL(\bq,\bp)$). We parameterize $\bq =
softmax(\bz)$ (with $T=1$) and we use gradient descent to optimize the logits $\bz$ w.r.t. eq.~\ref{eq:kl}. Note that this
optimization must be carried out for each image.  

\subsection{Results}

\begin{table}
\centering
\small
\begin{tabular}{|c|c|c|c|}
\hline
System & Conditional Test Accuracy & Test Accuracy \\
\hline
Baseline & 43.1\% & 25.0\%\\
 + 61 Specialist models & 45.9\%& 26.1\% \\
\hline
\end{tabular}
\caption{Classification accuracy (top 1) on the JFT development set.}\label{tab:image_results_spec}
\end{table}

Starting from the trained baseline full network,
the specialists train extremely fast (a few days instead of many weeks for JFT). Also, all the specialists
are trained completely independently.  Table ~\ref{tab:image_results_spec} shows the absolute test accuracy for the
baseline system and the baseline system combined with the specialist
models.  With 61 specialist models, there is a
4.4\% relative improvement in test accuracy overall. We also report conditional test accuracy, which is the accuracy by only considering examples belonging to the specialist classes, and restricting our predictions to that subset of classes.

\begin{table}
\centering
\small
\begin{tabular}{|r|r|r|r|}
\hline
\# of specialists covering & \# of test examples & delta in top1 correct &
relative accuracy change \\
\hline
  0 &  350037 &      0 & 0.0\% \\
  1 &  141993 &  +1421 & +3.4\% \\
  2 &   67161 &  +1572 & +7.4\% \\
  3 &   38801 &  +1124 & +8.8\% \\
  4 &   26298 &   +835 & +10.5\% \\
  5 &   16474 &   +561 & +11.1\% \\
  6 &   10682 &   +362 & +11.3\% \\
  7 &    7376 &   +232 & +12.8\% \\
  8 &    4703 &   +182 & +13.6\% \\
  9 &    4706 &   +208 & +16.6\% \\
 10 or more &    9082 &   +324 & +14.1\% \\
\hline
\end{tabular}
\caption{Top 1 accuracy improvement by \# of specialist models covering
correct class on the JFT test set.}\label{tab:jft_specialists_histogram}
\end{table}

For our JFT specialist experiments, we trained 61 specialist models, each with 300 classes (plus the dustbin class).
Because the sets of classes for the specialists are not disjoint, we often had multiple specialists covering a
particular image class.  Table ~\ref{tab:jft_specialists_histogram} shows the number of test set examples, the change in
the number of examples correct at position 1 when using the specialist(s), and the relative percentage improvement in
top1 accuracy for the JFT dataset broken down by the number of specialists covering the class.  We are encouraged by the
general trend that accuracy improvements are larger when we have more specialists covering a particular class, since
training independent specialist models is very easy to parallelize.

\section{Soft Targets as Regularizers}

One of our main claims about using soft targets instead of hard targets is that a lot of helpful information can be
carried in soft targets that could not possibly be encoded with a single hard target. In this section we demonstrate
that this is a very large effect by using far less data to fit the 85M
parameters of the baseline speech
model described earlier.  Table~\ref{tab:small} shows that with only 3\% of the data (about 20M examples), training the baseline model with
hard targets leads to severe overfitting (we did early stopping, as the accuracy drops sharply after reaching 44.5\%),
whereas the same model trained with soft targets is able to recover almost all the information in the full training set
(about 2\% shy). It is even more remarkable to note that we did not have to do early stopping: the system with soft
targets simply ``converged'' to 57\%. This shows that soft targets are a very effective way of communicating the
regularities discovered by a model trained on all of the data to another model.

\begin{table}
\centering
\small
\begin{tabular}{|l|c|c|}
\hline
System \& training set & Train Frame Accuracy & Test Frame Accuracy \\
\hline
Baseline (100\% of training set) & 63.4\% & 58.9\% \\
Baseline (3\% of training set) & 67.3\%& 44.5\% \\
Soft Targets (3\% of training set) & 65.4\%& 57.0\%\\
\hline
\end{tabular}
\caption{Soft targets allow a new model to generalize well from only 3\% of the
  training set. The soft targets are obtained by training on the full training set.}\label{tab:small}
\end{table}

\subsection{Using soft targets to prevent specialists from
  overfitting}

The specialists that we used in our experiments on the JFT dataset
collapsed all of their non-specialist classes into a single dustbin
class. If we allow specialists to have a full softmax over all
classes, there may be a much better way to prevent them overfitting than using
early stopping. A specialist is trained on data that is highly
enriched in its special classes.  This means that the effective size
of its training set is much smaller and it has a strong tendency to
overfit on its special classes. This problem cannot be solved by
making the specialist a lot smaller because then we lose the very
helpful transfer effects we get from modeling all of the
non-specialist classes. 

Our experiment using 3\% of the speech data strongly suggests that if
a specialist is initialized with the weights of the generalist, we can
make it retain nearly all of its knowledge about the non-special
classes by training it with soft targets for the non-special classes
in addition to training it with hard targets. The soft targets can
be provided by the generalist.  We are currently exploring this
approach.

\section{Relationship to Mixtures of Experts}

The use of specialists that are trained on subsets of the data has
some resemblance to mixtures of experts \cite{Jacobs} which use a
gating network to compute the probability of assigning each example to
each expert. At the same time as the experts are learning to deal with
the examples assigned to them, the gating network is learning to
choose which experts to assign each example to based on the relative
discriminative performance of the experts for that example.  Using the discriminative
performance of the experts to determine the learned assignments is much
better than simply clustering the input vectors and assigning an
expert to each cluster, but it makes the training hard to parallelize: First, the
weighted training set for each expert keeps changing in a way that
depends on all the other experts and second, the gating network needs
to compare the performance of different experts on the same example to
know how to revise its assignment probabilities.  These difficulties
have meant that mixtures of experts are rarely used in the regime
where they might be most beneficial: tasks with huge datasets that
contain distinctly different subsets.

It is much easier to parallelize the training of multiple specialists.
We first train a generalist model and then use the confusion matrix to
define the subsets that the specialists are trained on.  Once these
subsets have been defined the specialists can be trained entirely
independently.  At test time we can use the predictions from the
generalist model to decide which specialists are relevant and only
these specialists need to be run.

\section{Discussion}

We have shown that distilling works very well for transferring
knowledge from an ensemble or from a large highly regularized model
into a smaller, distilled model. On MNIST distillation works remarkably
well even when the transfer set that is used to train the distilled
model lacks any examples of one or more of the classes. For a deep
acoustic model that is version of the one used by Android voice search, we have shown
that nearly all of the improvement that is achieved by training an
ensemble of deep neural nets can be distilled into a single
neural net of the same size which is far easier to deploy.

For really big neural networks, it can be infeasible even to train a
full ensemble, but we have shown that the performance of a single really big
net that has been trained for a very long time can be significantly
improved by learning a large number of specialist nets, each of which
learns to discriminate between the classes in a highly confusable
cluster. We have not yet shown that we can distill the knowledge in
the specialists back into the single large net.

\subsubsection*{Acknowledgments}

We thank Yangqing Jia for assistance with training models on ImageNet and Ilya
Sutskever and Yoram Singer for helpful discussions. 

\bibliography{refs10}
\bibliographystyle{plain}

\end{document}